\documentclass{sig-alternate}

\usepackage{cite}
\usepackage{graphicx}
\usepackage{subcaption}
\usepackage[font=small]{caption}
\usepackage{float}
\usepackage{amsmath}
\usepackage{enumerate}
\usepackage{array}
\usepackage{url}

\begin{document}
%
\conferenceinfo{HuEvent'14}{November 07 2014, Orlando, FL, USA}
\CopyrightYear{2014} 
\crdata{978-1-4503-3120-3/14/11}  

\title{Investigating Human Factors in Image Forgery Detection}
%
%
%
%
%

\numberofauthors{2} 
%
\author{
%
%
\alignauthor
Parag Shridhar Chandakkar\\
       \affaddr{Computer Science and Engineering}\\
       \affaddr{Arizona State University}\\
       \email{pchandak@asu.edu}
\alignauthor
Baoxin Li\\
       \affaddr{Computer Science and Engineering}\\
       \affaddr{Arizona State University}\\
       \email{baoxin.li@asu.edu}
}
\date{07 April 2014}

\maketitle
\begin{abstract}
In today's age of internet and social media, one can find an enormous volume of forged images on-line. These images have been used in the past to convey falsified information and achieve harmful intentions. The spread and the effect of the social media only makes this problem more severe. While creating forged images has become easier due to software advancements, there is no automated algorithm which can reliably detect forgery.

Image forgery detection can be seen as a subset of image understanding problem. Human performance is still the gold-standard for these type of problems when compared to existing state-of-art automated algorithms. We conduct a subjective evaluation test with the aid of eye-tracker to investigate into human factors associated with this problem. We compare the performance of an automated algorithm and humans for forgery detection problem. We also develop an algorithm which uses the data from the evaluation test to predict the difficulty-level of an image\footnote{The difficulty-level of an image here denotes how difficult it is for humans to detect forgery in an image. Terms such as ``Easy/difficult image'' will be used in the same context.}. The experimental results presented in this paper should facilitate development of better algorithms in the future.
\end{abstract}




\begin{figure}[!t]
    \flushleft
    \begin{subfigure}[!t]{0.3\textwidth}
        \includegraphics[width=0.9\textwidth,height=70pt]{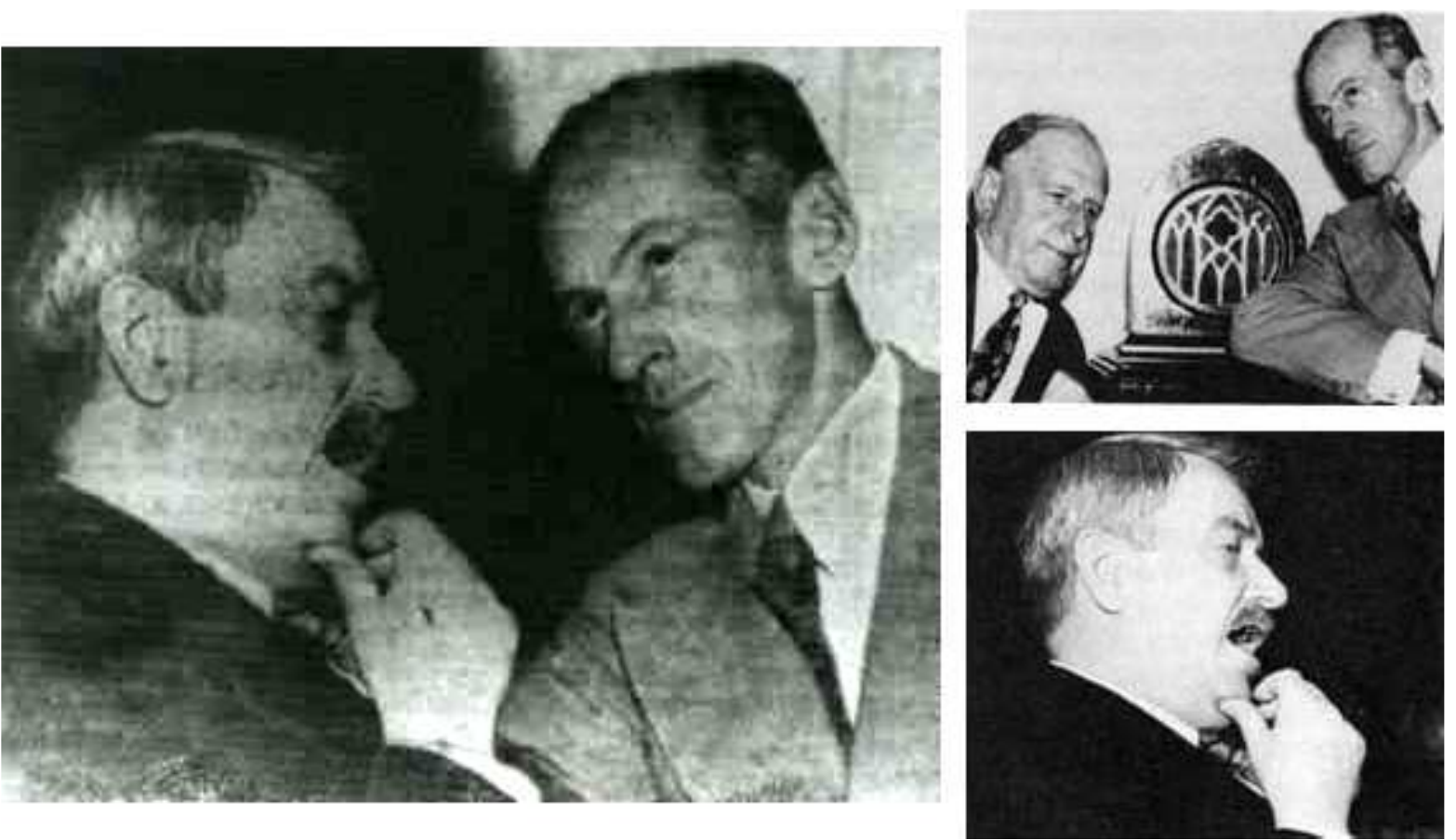}
        \caption{}
        \label{fig:Image 1}
    \end{subfigure}
    \hspace{6pt}
    \begin{subfigure}[!t]{0.15\textwidth}

        \includegraphics[width=0.9\textwidth,height=70pt]{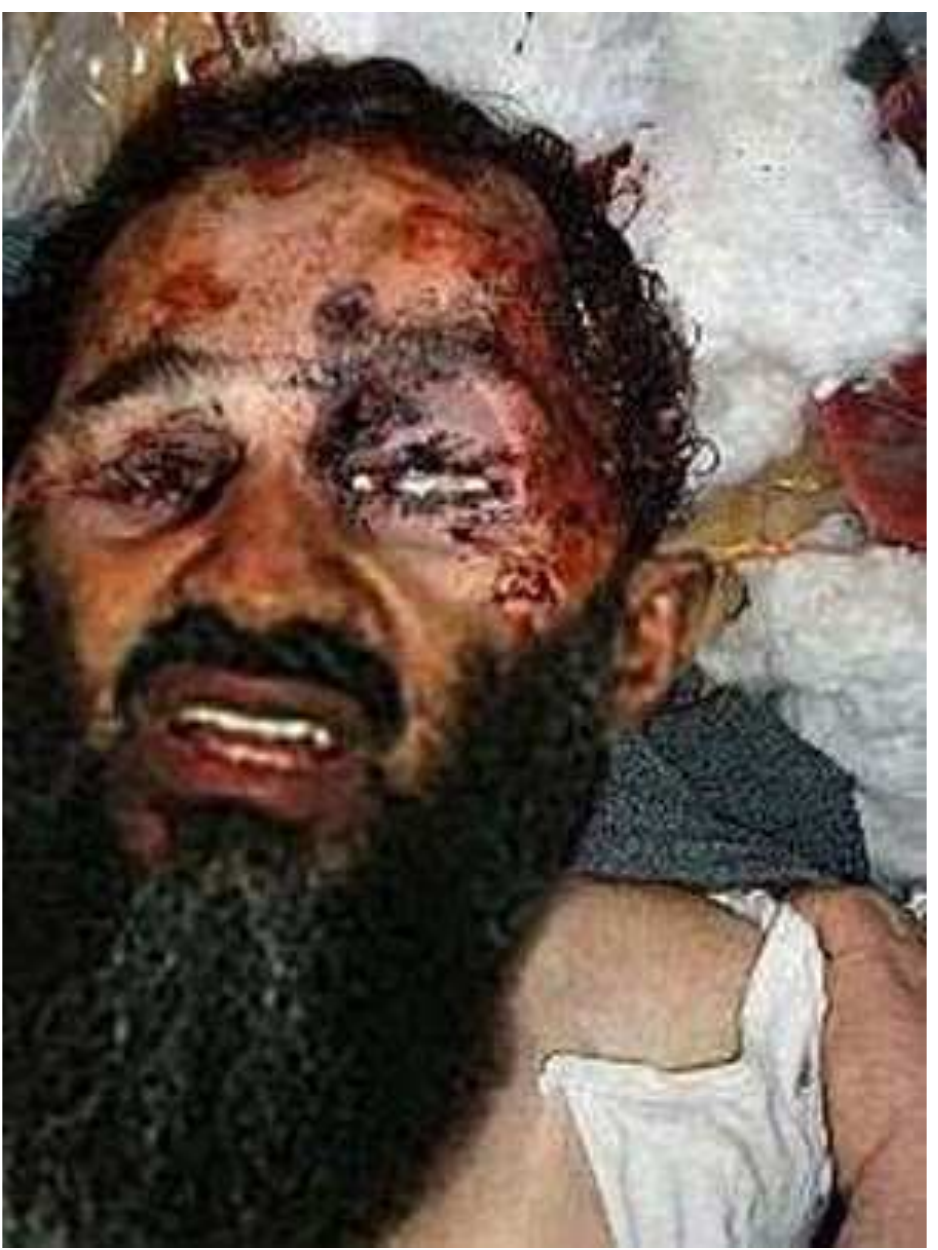}
        \caption{}
        \label{fig:Image 2}
    \end{subfigure}
    \caption{Examples of infamous tampered images.}
    \label{fig:Tampered images}
\end{figure}

\section{Introduction and related work}

Forged images are in abundance in today's age of social media and internet. They can be used to spread false information through social media and thereby achieve harmful intentions. They have been used in areas such as sports, fashion, politics, professional photography etc. for different motives. History of image forging dates back to 1800's, then mostly done for political reasons. We present two of the most infamous cases of forgery to show the severity of the problem. First infamous incident occurred in 1950 when a forged photo reportedly contributed to the electoral defeat of Senator Millard Tydings (right). The photo in Figure \ref{fig:Image 1} shows Millard Tydings having a conversation with Earl Browder (left), who was a leader of American Communist Party. In the second incident, Fig. \ref{fig:Image 2} shows a photo of Osama Bin Laden after his encounter with US forces on May $2^{nd}, 2011$. Though the photo was reportedly published in many places, it was later determined to be fake.

Image forgery can be categorized into two types: 1. Image splicing 2. Image tampering. Splicing is the simplest form of forgery where no post-processing is performed on the image. Tampering involves certain post-processing operations such as blurring, resizing etc. They are performed on the image to make it look as natural as possible. Most of the images on web are tampered. Creating forged images has become easier and detecting them is getting difficult due to constant advancements in editing software. There exist two common approaches to detect forgery, namely active and passive. Watermarking is an example of an active approach. It requires extra effort and most of the images on the web are not watermarked. Passive approach determines the authenticity of an image by analysing the image itself. The forgery detection process employed by humans is an example of a passive approach.

Image forgery detection is a subset of image understanding problems which include scene classification, object detection etc. Human performance is still the gold-standard for most of these problems. We investigate if the claim holds true for forgery detection. We conduct a subjective evaluation test with eye-tracking to quantify the human performance and to understand the behavioural aspect of this problem. In an attempt to relate human vision and the forgery detection process, we examine the relation between the saliency of an image and its effect on prediction performance. By using the eye-tracking data and the performance statistics from the evaluation test, we develop an algorithm to predict the difficulty-level of an image. We also compare the human performance against that of an automated algorithm. We envision the development of better algorithms in the future using these findings.

Recent years have seen an active research in this area. Copy-and-move forgery (CMF) is one of the most common methods of forgery in digital images. SURF-feature and textural-descriptors were used to detect CMF \cite{bo2010image,ardizzone2010copy}. Forgery in JPEG images is detected by analysing the DCT coefficients as the forged image is most likely to be compressed twice \cite{he2006detecting}. Another class of approaches uses high-level information in an image, such as, shadows \cite{kee2013exposing}, light environment \cite{ramamoorthi2001relationship} etc. Approaches solely depending on image statistics are image-format independent and are more computationally complex. Hilbert-Huang transform and Markov transition matrix of block DCT coefficients were proposed in \cite{fu2006detection} and \cite{sutthiwan2011markovian} respectively. We refer the reader to \cite{farid2009image} and \cite{mahdian2010bibliography} for an extensive review of forgery detection approaches.

This paper focuses on human performance evaluation of forgery detection. To the best of our knowledge, this work is first of its kind. Human performance evaluation studies have increased performance of object detectors and annotation predictors in \cite{yun2013studying}. Eye-tracking has also been used to study the behavioral aspects of radiologist's performance \cite{krupinski1996visual}.

The rest of the paper is structured as follows. Section 2 describes the proposed approach. The results and discussions are presented in section 3. Section 4 concludes the paper and lists potential future works.

\section{Proposed Approach} \label{sec: proposedApproach}
The aim of this paper is to examine human factors related to the forgery detection problem. The detection performance of many subjects on a set of forged images is recorded to present a quantitative assessment of the human performance. We analyse the human gaze-points with the aid of eye tracker to understand the behavioural aspect. It was shown in \cite{yun2013studying} that there is a strong relation between the human-level understanding of an image and the human gaze-points. The experimental set-up and protocol is designed to establish a relationship between the pattern of human gaze-points and the difficulty-level of a forged image.

\subsection{Dataset and Experimental set-up} \label{subsec: experiment}
We present a quantitative measure of human performance and study the relationships between the human gaze and image content using images from two standard image forgery datasets, CASIA v1.0 and v2.0 (from http://forensics.\newline idealtest.org). CASIA v1.0 is a splicing dataset whereas its next version has tampered images. Our database has 73 images out of which 14 are spliced (taken from CASIA v1.0), 44 are tampered (taken from CASIA v2.0) and 15 are authentic images. The evaluation test is performed with 24 subjects over a span of one week. Subjects are divided into 3 groups of 9, 9 and 6. Each subject in the first two groups is presented with 50 images at a time. Subjects in third group are presented with all 73 images. The images stay the same within a group and those 50 images are sampled from 73 images. All the subjects sat at an approximate distance of 25 inches from a 15.6 inch screen having resolution of $1920 \times 1080$. Natural lighting conditions are used for the test and no time-limit is imposed to allow the subjects to scan the image at their own pace. The distribution of authentic and forged images is unknown to subjects. Before starting the evaluation, five examples of each kind of images are shown to each subject with the forged region being pointed out in order to give them an idea about the type of forgery. They are instructed to classify each image into one of two groups, namely, authentic or forged and point out the forged region.

\begin{figure}[!t]
        \includegraphics[width=0.5\textwidth,height=119pt]{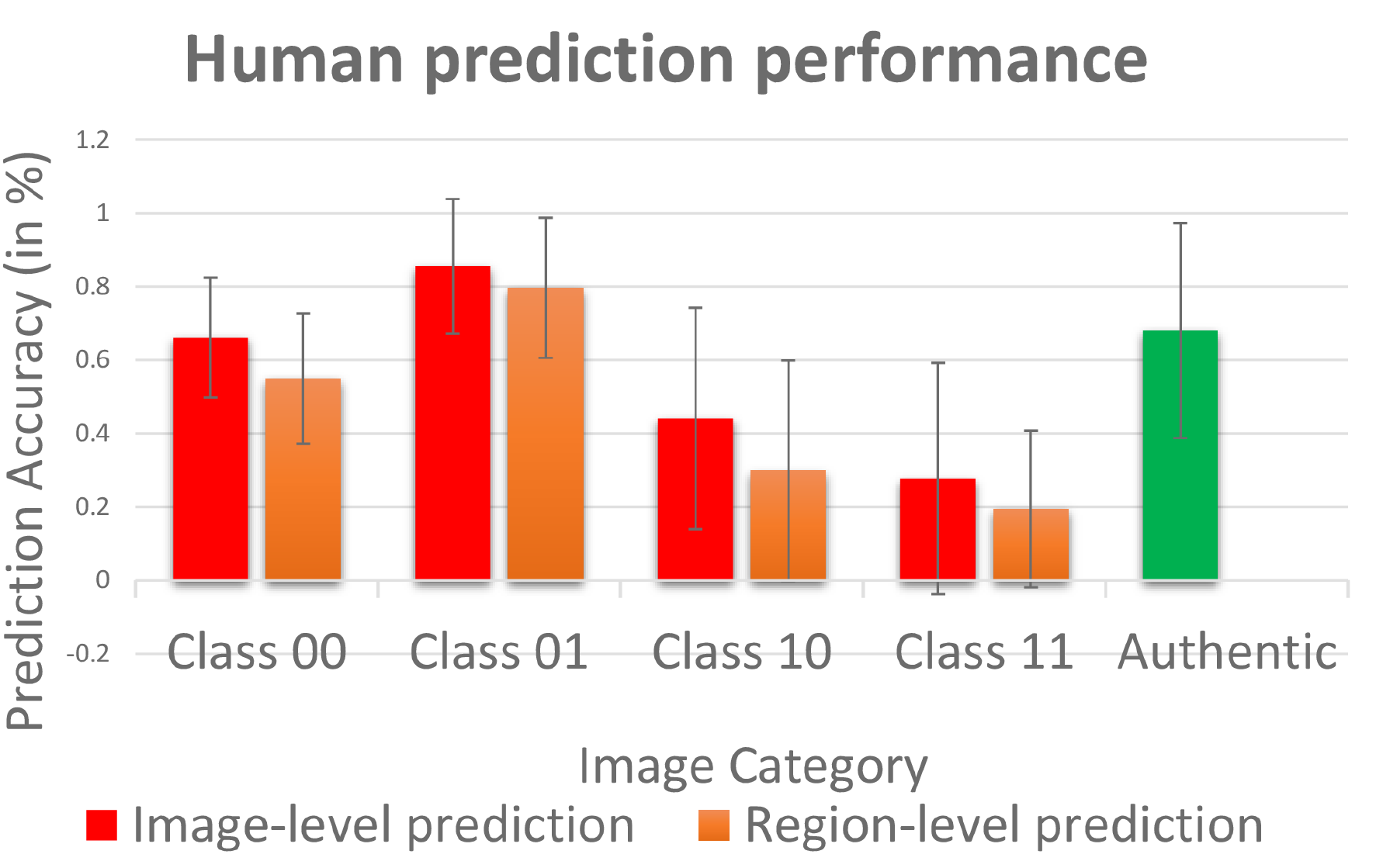}
        \caption{Human prediction performance grouped by category}
        \label{fig:detAccPerClass}
\end{figure}

\subsection{Analysis} \label{subsec: analysis}
The eye-tracker data and image-level statistics are used to address several questions relating gaze and other factors which may affect the difficulty-level of an image. The contribution of our analysis is three-fold. Firstly, we provide a quantitative measure of human performance on our dataset. We use the model of image saliency to predict the difficulty-level of an image. Third contribution aims at developing an automated algorithm to predict the difficulty-level using data from saliency model and gaze-points.

\subsubsection{Relating image saliency with forgery}  \label{subsubsec: saliencyAnalysis}
For every forged image in the database, we have their authentic counterparts available for analysis. We propose to model the change in saliency between the forged image and its authentic counterpart. We compute saliency in spatial \cite{margolin2013makes} as well as frequency-domain \cite{hou2012image}. Depending on the way an image has been forged, we categorize an image into one of four classes as follows:
\vspace{-2pt}
\begin{itemize}
\item Class 01: The forged part is \underline{non-salient} in the authentic image and becomes \underline{salient} in the forged image.
\end{itemize}
\vspace{-2pt}
Class 00, 10 and 11 can be similarly described. We find that change in saliency significantly affects the difficulty-level of an image. Fig. \ref{fig:detAccPerClass} shows the category-wise accuracy. Quantitative results are given in section \ref{sec:results}.

\subsubsection{Relating gaze with forgery}  \label{subsubsec: gazeAnalysis}
Human gaze contains abundant information about the task and human thought-process \cite{yarbus1967eye}. We analyse the information and come up with the following metrics to better explain the human process of forgery detection.

\begin{figure}[!t]
\centering
        \includegraphics[width=0.45\textwidth,height=125pt]{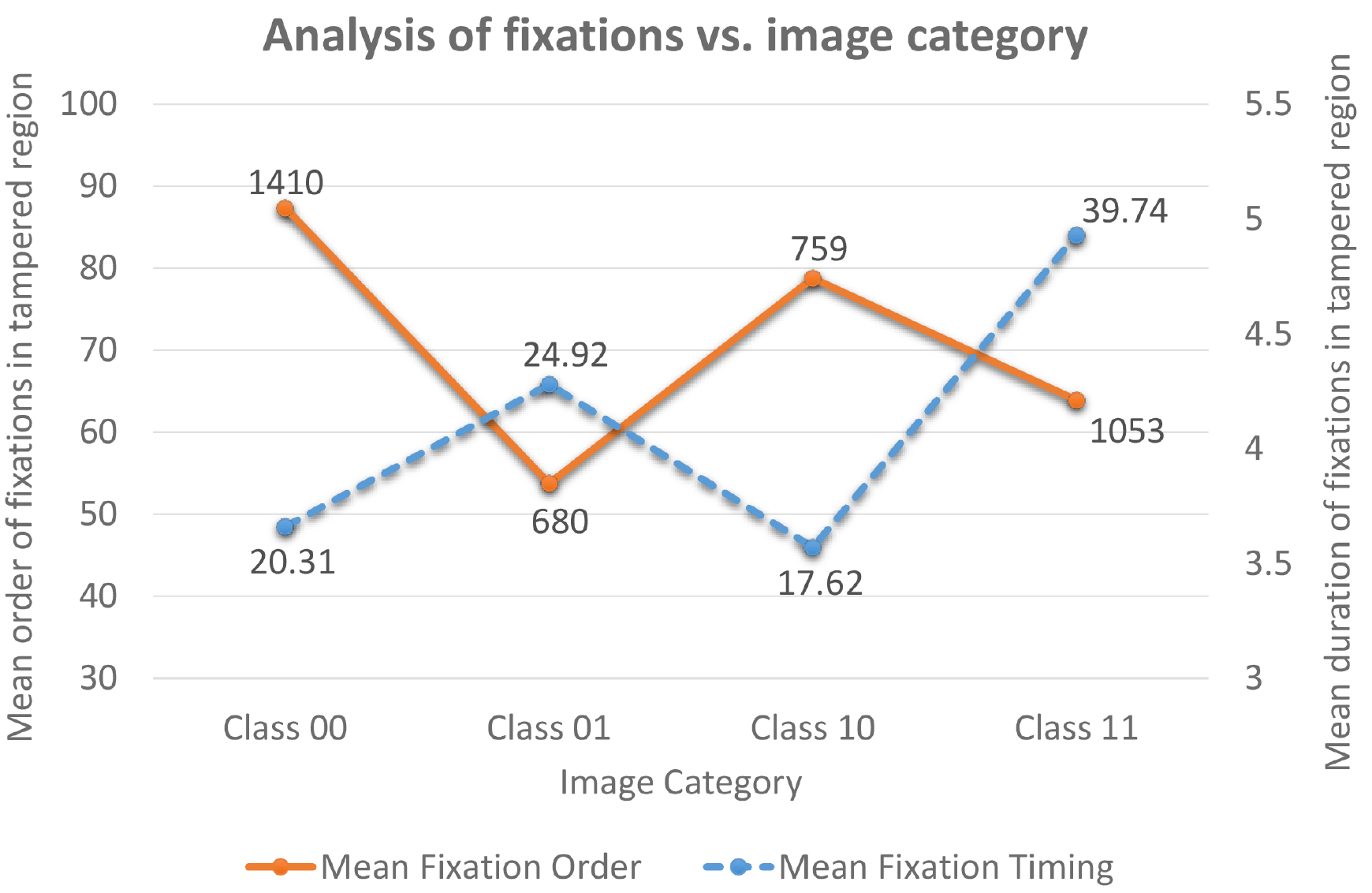}
        \caption{Effect of image category on fixations}
        \label{fig:fixnPerClass}
\end{figure}

\begin{enumerate}
\item For class 00 and 10, we compute the following metric:

\begin{equation}
\text{Gaze-metric1} = \frac{\# \text{ fixations in forged region}}{\# \text{ fixations in salient region}}
\end{equation}

Salient region in an image can be detected by any standard saliency algorithm. We define \textit{Gaze-metric2} for class 01 and 11. We count the number of fixations lying elsewhere instead of in salient region since salient and forged region intersect. This gives a measure about the number of fixations used to detect forgery. Intuitively, \textit{Gaze-metric1} (\textit{Gaze-metric2}) should get higher values for easy images. Higher values for both metrics imply shorter fixation duration in the salient region (elsewhere) than in the forged region. Since human vision is usually clustered in the salient part of an image, this behaviour supports the claim of task-driven human vision \cite{yarbus1967eye}.\\[-5mm]

\item We analyse the effect of image category on duration of fixations in the forged region. For each category, we take average of the duration of fixations in the forged regions. Note that we average over images in a particular category as well as over all the subjects. We also record the order of fixations in the forged regions while subjects scan the given image. The plot of mean order of fixations versus image category is shown in Fig. \ref{fig:fixnPerClass}. The analysis of fixation statistics is given in section \ref{sec:results}.\\[-5mm]

\item We study the effect of fixation duration over an entire image on the prediction accuracy. We add-up duration of fixations for all the images for each subject. Scanning an image for a longer duration can be related to analysing image over multiple scales and regions. The plot is shown in Fig. \ref{fig:fixnDurn}.\\[-5mm]

\end{enumerate}

\noindent Finally, we use clustering to group the features from the gaze-data and saliency model into two clusters. The images in a cluster should correspond to the same category, i.e., easy or difficult. Thus a large difference between the accuracies of two clusters (averaged over all the subjects) is expected.

\section{Results and discussions} \label{sec:results}
In this section, we present the quantitative results and discuss the findings for various metrics presented in section \ref{subsec: analysis}. Human prediction performance on authentic as well as forged images is shown in Fig. \ref{fig:detAccPerClass}. The green and red bars represent prediction performance on an image-level for authentic and forged images respectively. The orange bars represent region-level prediction accuracy for forged images. The error bars represent the standard deviation in a particular category (calculated over all the subjects). 

\subsection{Saliency analysis}
The Highest performance is obtained in class 01 as expected due to saliency of forged region. The lower performance associated with class 10 can be associated to the non-saliency of the forged region after it undergoes editing. The forgery operation in the 10 category usually involves removal of the salient region in the authentic image and replacing it with a non-salient texture, which is un-noticed by many subjects. However, lowest accuracy is achieved on the class 11, indicating that saliency is not necessarily a factor which decides difficulty of a forged image. Images in the class 11 are skillfully forged by preserving the spatial and contextual continuity after editing. This makes forgery detection difficult for superficial observers. Second-highest performance is obtained on class 00 which further supports our claim about saliency not being a deciding factor.

\begin{figure}[!b]
        \includegraphics[width=0.485\textwidth,height=130pt]{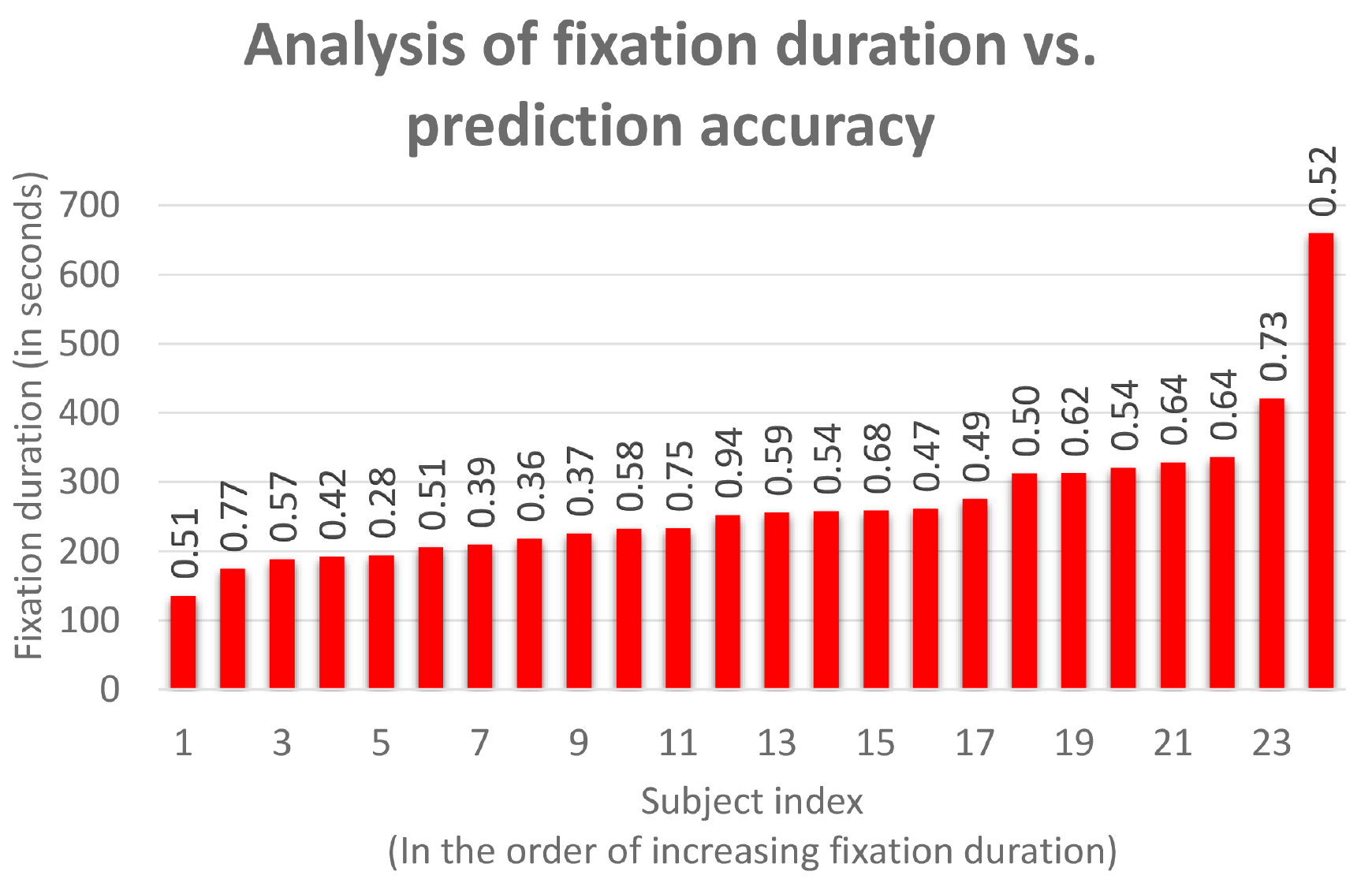}
        \caption{Effect of fixation duration on prediction.}
        \label{fig:fixnDurn}
\end{figure}

\subsection{Fixation analysis}
Two gaze metrics proposed in the section \ref{subsubsec: gazeAnalysis} can be successfully used to predict easy images. Values of both the metrics are collected for all the images in respective classes and are sorted. The values are binary thresholded at a point where large change is observed and are grouped into 2 groups per metric. Group 1 and 3 (2 and 4) contains metric values above (below) threshold. The thresholds are determined to be 4 and 2 respectively for two metrics. Accuracies of all the images are also classified into 2 groups as per the obtained thresholds. The mean accuracies of images (averaged over all subjects) in the 4 groups are found to be $75\%$, $58.7\%$, $66.7\%$ and $47.8\%$ respectively in accordance with our analysis.

\begin{figure}[!t]
    \flushleft
    \begin{subfigure}[!t]{0.22\textwidth}
        \includegraphics[width=\textwidth]{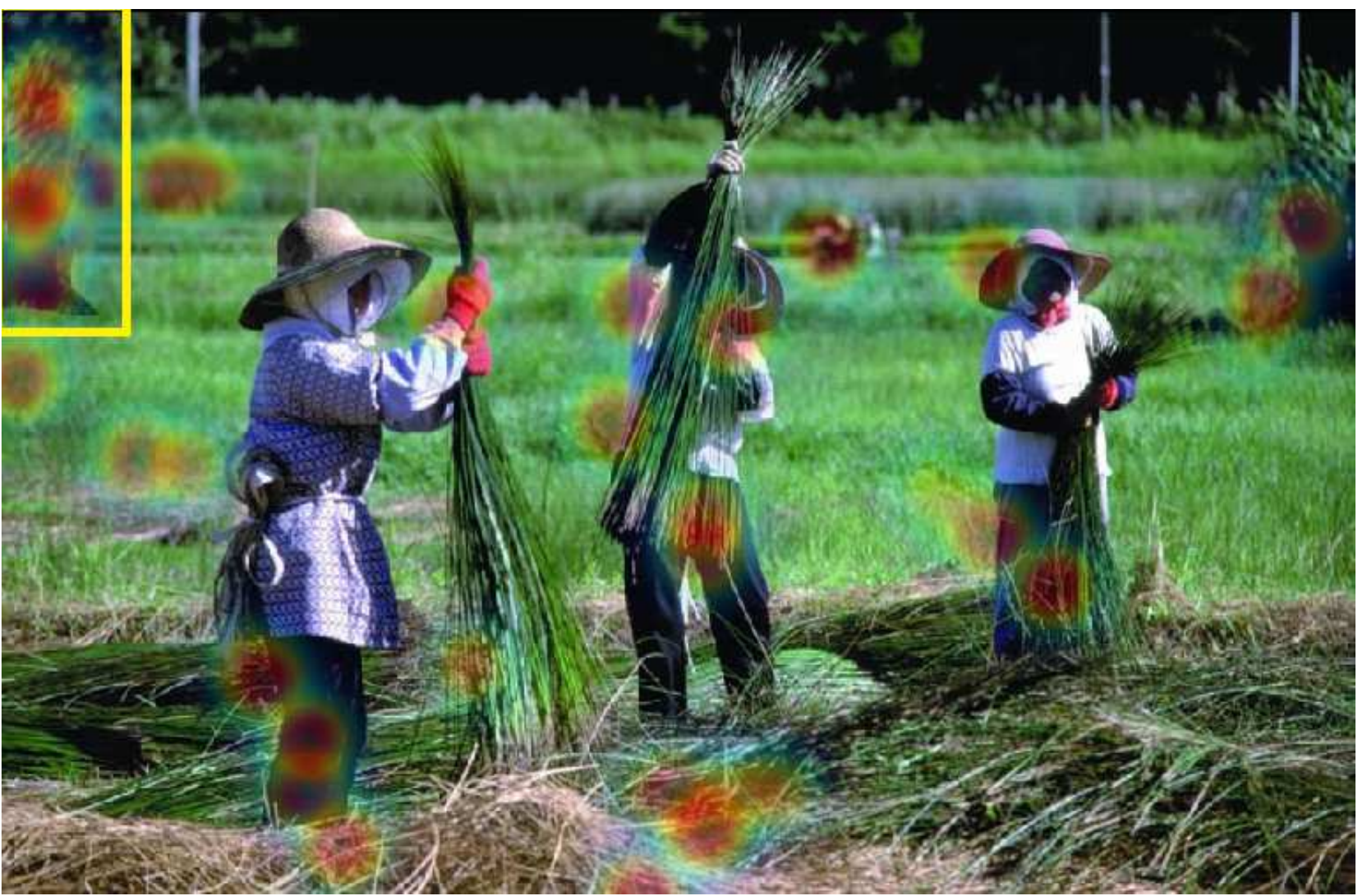}
        \caption{}
        \label{fig:Heatmap 1}
    \end{subfigure}
    \hspace{6pt}
    \begin{subfigure}[!t]{0.22\textwidth}

        \includegraphics[width=\textwidth,height=80pt]{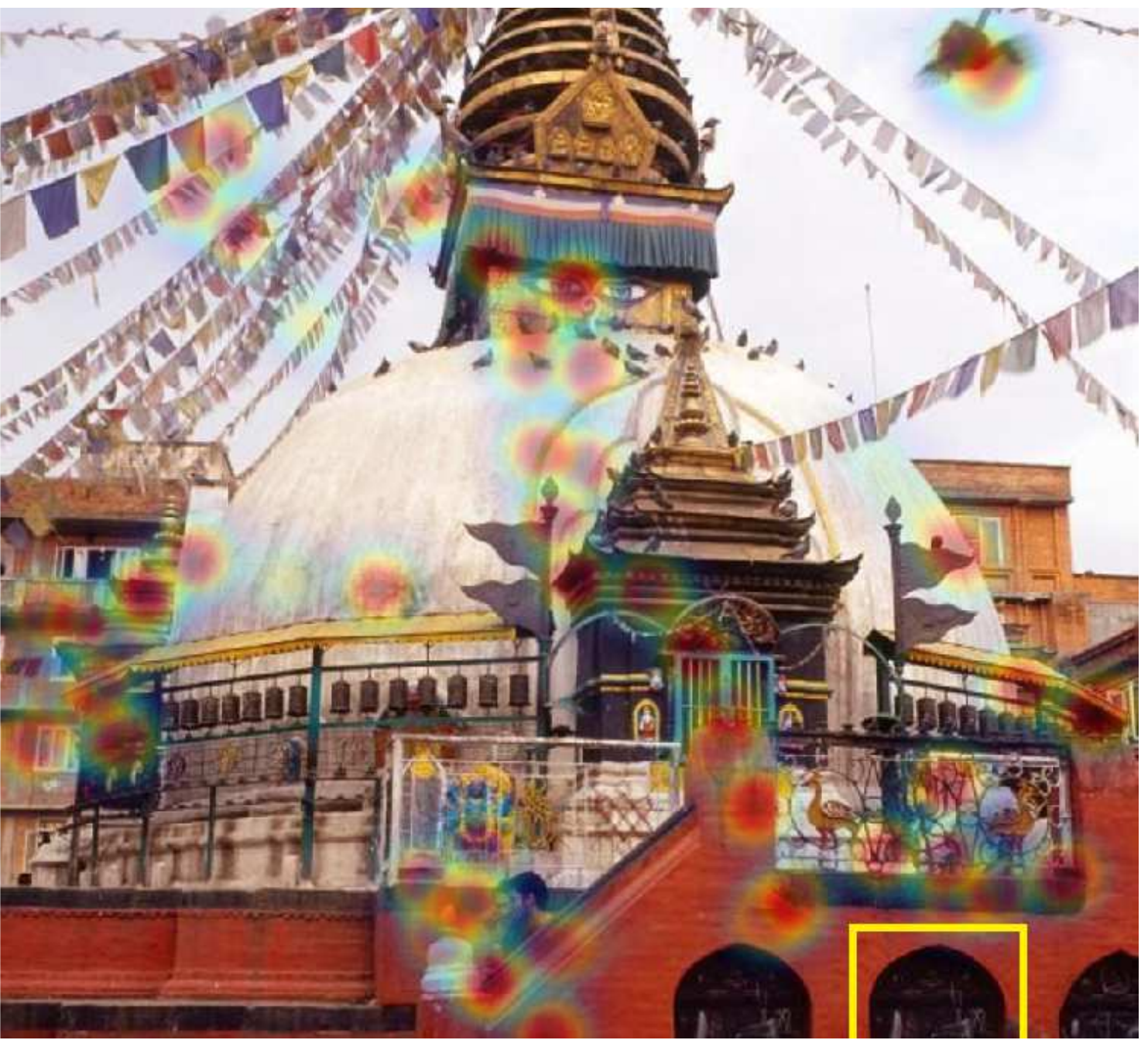}
        \caption{}
        \label{fig:Heatmap 2}
    \end{subfigure}
    \caption{Fixation heatmap of (a) correct detection (b) missed detection of forged images. Forged area is highlighted in yellow (Best viewed in color).}
    \label{fig:Heatmaps}
\end{figure}

Next, we analyse the effect of image category on fixation duration and order. The data-labels plotted on top of each data-point in Fig. \ref{fig:fixnPerClass} represent the maximum value. For example, $1410$ denotes the maximum (over all subjects) number of fixations required to fixate a subject's vision on forged region in all the images in class 00. Similarly, $20.31$ denotes the maximum amount of time a subject fixated in forged region. This plot shows that fixation duration and order are both affected by the saliency. For class 01, subjects fixated on the forged region earlier whereas most number of fixations were required in the class 00. It is interesting that in spite of the contrast in the fixations, these two categories get the top-2 prediction scores. In class 10, subjects fixated on the forged region for the least amount of time. On the other hand, they fixated for the most amount of time in class 11, probably because the forged region also intersects with the salient region. Yet, these two classes have two lowest prediction scores associated with them. Human fixations are thus affected by the image saliency and the given task.

What would be the effect of fixating longer at an image? Will it result in increased prediction accuracy? The plot in Fig. \ref{fig:fixnDurn} suggests otherwise. In the plot, data-label over each data-point denotes the accuracy obtained by the subject. The subjects are arranged in the increasing order of fixation duration. The pattern suggests that fixation duration over an entire image contributes little to the prediction process.

\subsection{Comparison with automated algorithm}
We implement a passive approach for image forgery detection proposed in \cite{sutthiwan2011markovian}. On the subset of 50 images, best human and computer performances are found to be $72\%$ and $62\%$ respectively. For all 73 images, the same numbers are $68.49\%$ and $68.49\%$ respectively. We also examine the class-wise performance. Best Human performance on the 4 classes and on authentic set is $75\%$, $100\%$, $63.63\%$, $22.22\%$ and $60\%$ respectively. The same set of numbers for computer performance is $75\%$, $78.57\%$, $45.45\%$, $77.78\%$ and $60\%$ respectively. Thus humans are better at detecting unskilled forgery but on the difficult images automated algorithm performs slightly better.

\subsection{Prediction algorithm}

We develop an automated algorithm to group the images under two categories, namely, easy and difficult, by using the data from saliency and gaze-analysis . We use \textit{K}-means where the feature vector contains the following quantities: 1. Class index 2. Gaze metrics 3. Order of fixation in forged region 4. Duration of fixations in forged region.

After clustering the feature space into 2 groups, the accuracies for both the clusters obtained are $63.87\%$ and $37.69\%$. Note that the accuracies are averaged over all the images in that cluster and over all the users. In spite of averaging over all subjects and images, the collected features are able to fairly predict the difficulty level of an image.

\section{Conclusion and future work}

We investigate the human factors associated with the process of forgery detection by conducting a subjective evaluation test with the aid of eye-tracker. To analyze the effect of saliency on forgery detection, we group the images in four classes and perform statistical analysis. We show that though saliency affects the prediction, it is not necessarily a deciding factor. We use eye-tracker data and show the effect of image category on the duration of fixations and their order. We successfully apply clustering on the generated data to group the images into easy and difficult categories. We compare the performance of automated algorithm and humans to show that automated algorithm is better at detecting skilled forgery. In the future, we would like to closely examine the dependency between underlying statistics in the fixation pattern, image saliency and image forgery.

\textbf{Acknowledgement:} The work was supported in part by a grant (\#1135616) from the National Science Foundation. Any opinions expressed in this material are those of the authors and do not necessarily reflect the views of the NSF.

\bibliographystyle{abbrv}
\bibliography{sigproc}  
%
%
\end{document}